\documentclass{article}
\usepackage{ijcai26}

\usepackage{times}
\usepackage{soul}
\usepackage{url}
\usepackage[hidelinks]{hyperref}
\usepackage[utf8]{inputenc}
\usepackage[small]{caption}
\usepackage{graphicx}
\usepackage{amsmath}
\usepackage{amsthm}
\usepackage{booktabs}
\usepackage{algorithm}
\usepackage{algorithmic}
\usepackage{amsfonts}
\usepackage{mathtools}
\usepackage{multirow}
\usepackage{adjustbox}
\usepackage{array}
\usepackage{xcolor}
\usepackage{textcomp}
\usepackage{listings}
\usepackage[switch]{lineno}

\usepackage[title,titletoc,page]{appendix}

\title{What Would an LLM Do? Evaluating Large Language Models for Policymaking to Alleviate Homelessness}

\author{
Pierre Le Coz$^1$\And
Jia An Liu$^1$\And
Debarun Bhattacharjya$^2$\And
Georgina Curto$^1$\And
Serge Stinckwich$^1$\\
\affiliations
$^1$United Nations University Institute in Macau, Macau SAR, China\\
$^2$IBM Research, Yorktown Heights, NY, USA\\
\emails
lecoz@unu.edu, jiaan@unu.edu, debarunb@us.ibm.com, curto@unu.edu, stinckwich@unu.edu
}
\begin{document}

\maketitle

\begin{abstract}

Large language models (LLMs) are increasingly being
adopted in high-stakes domains. Their potential to encode evolving social contexts and to generate plausible scenarios position them as promising tools in social policymaking. This article evaluates whether
LLMs are aligned with domain experts (and among themselves) on policy recommendations to alleviate homelessness – a challenge affecting over 150 million
people worldwide. 
We develop a novel benchmark comprised
of decision scenarios across four cities, with policy choices that are grounded in the conceptual framework of the Capability
Approach for human development. 
We also present an automated pipeline that connects the  policies to an agent-based model in one location, and compare the social impact of the policies recommended by LLMs to those recommended by experts. Our exploratory analysis reveals variation across LLMs in their policy recommendations compared to local experts, yet suggests potential benefits of the use of LLMs to provide insights for policymaking, if paired with responsible guardrails, contextual calibrations, and local domain expertise. Our work operationalizes the Capability Approach in a computational framework and provides new insights on homelessness alleviation policymaking with a focus on human dignity.  

\end{abstract}


\section{Introduction}

Homelessness, defined by the United Nations (UN) as ``the lack of a stable, safe, and adequate housing'' \cite{ohchr_homelessness}, is a growing crisis affecting cities and nations worldwide. 
According to the UN, an estimated 1.6 billion people globally lack adequate housing, with 150 million completely homeless. The OECD reports that in most developed countries, homelessness has sharply increased since 2022 \cite{OECD2024_HM1.1}, with the U.S. alone recording more than 653,000 unhoused individuals in a single night in January 2023 \cite{StateofhomelessnessUSA}.
Homelessness represents a severe deprivation of basic human security and dignity, undermining sustainable development goals, 
and tackling these challenges constitutes a core requirement for creating equitable, resilient societies where no one is left behind \cite{UN2030Agenda}.

In response to these urgent social challenges, policymaking to alleviate homelessness needs to extend beyond material redistribution to confront a more profound requirement: ensuring unhoused individuals are treated as equals and fully integrated into the social fabric. 
This necessitates dismantling structural stigmatization and fostering genuine belonging -- a task complicated by deeply entrenched societal biases, bureaucratic systems that often reduce people to ``cases'', and the trauma-induced isolation experienced by those without stable shelter \cite{Curto2025,Ranjit2024}.

Effective homelessness mitigation policies must not only allocate material resources but actively combat dehumanization and societal rejection \cite{takahashi1998homelessness,WorldBank2000,WassermanClair2009}, centering the agency and dignity of those affected, and recognize inclusion as both an ethical imperative and a pragmatic necessity. 
Given the multifaceted nature of homelessness and the difficulty in anticipating policy outcomes in dynamic social systems, computational approaches offer a promising avenue to navigate this complexity and enable prospective policy testing. Recent advances in computational social science have demonstrated the potential of agent-based models (ABMs) to inform homelessness policymaking, representing homelessness as a multidimensional deprivation \cite{Aguilera2025_ECAI}. However, significant challenges remain in modeling human behavior within social simulations, such as the complexity to capture latent psychological factors and social values ~\cite{dignumSocialSimulationCrisis2021}, as well as the difficulty around developing generalizable models that align with different local policies, evolving social contexts, and resource availability across geographies~\cite{Aguilera2025_ECAI} 

In this context, the increasingly widespread integration of LLMs in societal systems, their capacity to process vast amounts of unstructured data, explore flexible role-play scenarios, and handle a diversity of evolving contextual factors can make them uniquely suited to provide new insights for decision making in complex scenarios \cite{Wang2024,Gao2023,Guo2024}. 
Despite these capabilities, deploying LLMs in socially sensitive domains, such as homelessness policymaking, presents severe challenges. These include the amplification of societal biases \cite{Agnew2024}, generation of plausible but misleading outputs (“hallucinations”), a lack of genuine understanding of human suffering and social dynamics \cite{shanahanTalkingLargeLanguage2024}, and inherent opacity that makes their reasoning difficult to audit. It is therefore imperative to proactively assess their risks and limitations in such high-stakes contexts. 

This paper aims to address these limitations by 
comparing  recommendations made by LLMs with those provided by domain experts. We introduce a benchmark with decision scenarios that contain policy choices to reduce homelessness 
in four different parts of the world (South Bend Indiana, USA; Barcelona, Spain; Johannesburg, South Africa; Macau SAR, China). 
The selection of policies included in the benchmark is grounded in the conceptual framework of the Capability Approach for human development (CA) \cite{sen1999,nussbaum2011,robeyns2017}. Unlike models that systematically prioritize the material needs of individuals \cite{maslow1943}, the CA considers the opportunities that human beings have to lead a meaningful life with dignity \cite{sen1999}. From this perspective, homelessness can be considered as a deprivation of central capabilities~\cite{nussbaum2011}. In our benchmark, the range of policies offered in the different scenarios illustrate the restoration of CA's central capabilities for population in a situation of homelessness. 
We perform an empirical evaluation around LLMs' choices and judgments on the benchmark, comparing them with those offered by the domain experts in each geographical location. 

Then, we propose a novel pipeline to automatically link policies  with an ABM framework, intended to gauge the social impacts of LLM generated policies, versus those proposed by domain experts, in a simulated social context. Specifically, this approach explores how policies inform agent behavior through an existing State-Action-Transition process (SAT)  \cite{Aguilera2024}. 
Our work opens up new directions to leverage LLMs for supporting social policymaking in a manner that is scalable and non-invasive, where LLMs could be used to suggest alternative policies aligned with the conceptual framework of the CA, and their impact could be examined via ABM simulations. The paper operationalizes the Capability Approach for human development in a computational framework, suggesting policies that go beyond covering the basic material needs of the persons with no home, with a focus on the whole social fabric and human dignity. The work is requested and conducted in close collaboration with domain experts and non-profit organizations specialized in alleviating homelessness in the four cities in scope. The domain experts participated in the definition of the scope and research approach. Multi-disciplinary domain experts defined the baseline scenario and conducted policy rankings in the benchmark scenarios. 
Insights from our work are being leveraged to inform
ongoing policymaking discussions 
in the four locations in scope.

  

\section{Related Work}

\subsection{The Capability Approach 
}

Sen and Nussbaum's work on the Capability Approach 
constitutes a cornerstone in human development studies and as a conceptual framework underpinning the Sustainable Development Goals. And its operationalization in the practice of social policy making still has much potential to unveil \cite{robeyns2017}. From a computational social practice perspective, only recently has the CA been integrated into 
methods such as agent-based modeling \cite{Aguilera2025_ECAI,Chavez-Juarez02012021}
To the best of our knowledge, our work constitutes the first application of CA as a framework for generative AI in social policy.

\subsection{Evaluating LLMs for Social Policymaking}

Recent research has explored the potential and limitations of LLMs in policymaking contexts. \citeauthor{jiao2025navigatingllmethicsadvancements} (\citeyear{jiao2025navigatingllmethicsadvancements}) systematically documented risks when deploying LLMs in high-stakes policy domains, highlighting issues of bias amplification and context blindness, which this paper aims to address.  Work by \citeauthor{WEI2025104103} (\citeyear{WEI2025104103}) provided evidence of how LLMs can perpetuate societal biases in resource allocation scenarios, raising critical questions about their suitability for welfare policymaking. Recurrent hallucinations in general problem solving~\cite{xu2025hallucinationinevitableinnatelimitation} takes on particular significance in homelessness policymaking, where factual inaccuracies could directly impact the most vulnerable individuals.

While frameworks exist for evaluating LLM outputs in healthcare decision making \cite{kanithi2024mediccomprehensiveframeworkevaluating} and in legal domains \cite{li2024legalagentbench}, homelessness alleviation policy presents unique challenges due to its intersection with complex socioeconomic factors and social stigmatization, which have different nuisances across geographies. 
Our work presents a novel framework to specifically assess the ability of LLMs to inform policy making on the topic of homelessness alleviation in four distinct geographic locations.



\subsection{Integrating LLMs with Agent-Based Social Simulations}

The integration of LLMs with ABMs in social simulations constitutes an emerging topic in computational social science \cite{Chopra2024,Guo2024,gaoLargeLanguageModels2024,Jin2025}. Traditional ABMs have proven to be very helpful to inform social policy making in a non-invasive manner \cite{dignumSocialSimulationCrisis2021}. 
LLMs have shown the potential to enrich and diversify the behavior of agents with reasoning-like abilities, memory, personality traits, and context sensitivity \cite{ferraro2024agentbasedmodellingmeetsgenerative,Park2024,Horto2023}. Although conceptual frameworks have been presented to connect LLMs with ABMs in socially sensitive frameworks \cite{Ricci2024}, the technical 
complexities remain a challenge and there is still a question about the reliability of recommendations provided by LLMs as inputs for policymaking. In this work, we present a pipeline that connects the policies recommended by LLMs, and we evaluate their impact on homelessness alleviation in an ABM framework. 

\section{Methodology}

\begin{figure*}[htb]
\centering
\includegraphics[width=\linewidth]{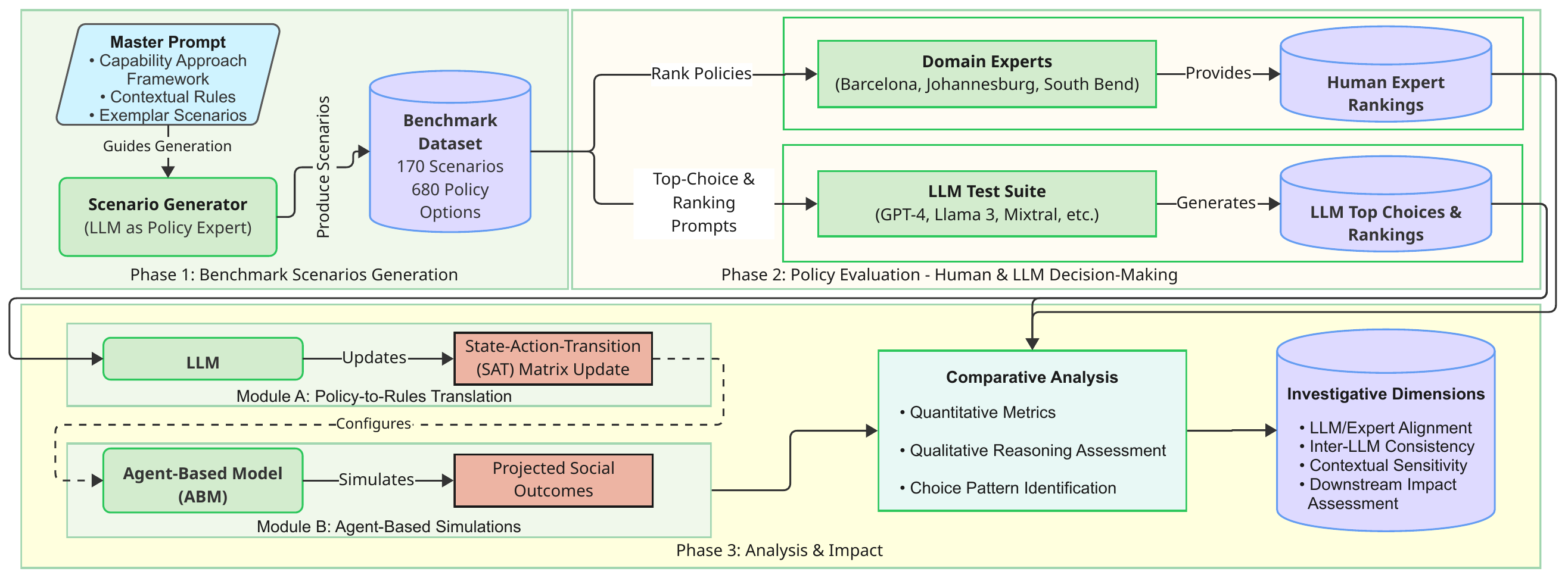}
\caption{Methodology overview: We construct a benchmark via a structured 
prompting strategy grounded in the Capability Approach for human development (CA) prompt various LLMs to act as policymakers,  analyze their choices through comparison with human expert recommendations, and conduct an evaluation of their projected societal impact using a modular agent-based modeling pipeline.  
}
\label{fig:comparing_LLMs_architecture}
\end{figure*}

In this section, we describe our methodology, an overview of which is illustrated in Figure~\ref{fig:comparing_LLMs_architecture}.


\subsection{Benchmark Design}

We present the components of a novel benchmark that aims to evaluate whether LLMs are aligned with domain experts working with specialized organizations, within the locations in scope, to inform homelessness alleviation policymaking\footnote{We will make the benchmark publicly available.}. 

\subsubsection{Dataset Composition}
\label{subsec:dataset-composition}

\paragraph{Locations and scope.}
The benchmark covers four geographic contexts: \textbf{Barcelona} (Spain), \textbf{Johannesburg} (South Africa), \textbf{South Bend} (USA), \textbf{Macau SAR} (China), as well as a \textbf{Universal} context not tied to any particular city. The universal context is included to study the impact of context on choices. For each of the four specific locations, we generate 40 decision-making scenarios, while the universal category contains 10 scenarios. Each scenario frames a dilemma faced by a non--profit organization  working to alleviate homelessness, e.g., how to allocate emergency housing funds, whether to prioritize medical outreach or job training, etc. 
In total, the benchmark comprises 170 scenarios, each with 4 choices. 

\paragraph{Scenario structure.}
Every scenario is presented as a structured JSON object with four fields: \textit{Scenario} (a title summarising the dilemma), \textit{Context} (one or more paragraphs providing background), \textit{Policy\_Options} (an array of four policy proposals) and \textit{Main\_capability\_restoration} (a list of central capabilities 
targeted by each policy). The context is written in narrative form and spans at least 80 words, capturing locally relevant demographics, political debates, economic constraints and historical factors. 
Each policy option is described in a paragraph of at least 35 words and is annotated with one or more of Martha Nussbaum’s 
ten central human capabilities (see Appendix A)\footnote{Appendices are in the supplementary material.}, 
which are universal freedoms that societies ought to protect -- ranging from the ability to live a full life and access healthcare to the freedom to develop practical reason and to participate in political decision making~\cite{Nussbaum2012}.
Using this normative framework ensures that each policy speaks to a specific aspect of human flourishing rather than merely enumerating material resources.

\subsubsection{Scenario Generation}
\label{subsec:scenario-generation}

\paragraph{Baseline scenario design.}
To seed the generation process, we first manually crafted a baseline scenario for Barcelona in consultation with domain experts. This scenario, rooted in current debates about housing  in the city \cite{diagnosi2022barcelona}, presented a homelessness dilemma alongside four plausible policy responses aiming to restore affected central capabilities. The manual scenario served both as a template for automated generation and as a qualitative anchor for calibrating LLM outputs.

\paragraph{Prompt-based generation.}
Subsequent scenarios were generated using a frontier LLM (GPT--4.1) instructed to act as a public policy expert with knowledge on urban development and the Capability Approach. We provided the LLM with a detailed project description, a list of the ten central capabilities, guidance on scenario length and tone, and examples of desirable output.
The model was prompted to produce exactly 40 non--redundant scenarios for each city (and 10 for the universal category). Each prompt emphasized that scenarios must (1) present a localized public policy dilemma related to homelessness, (2) be either grounded in a documented real situation or be a plausible fictional event, (3) include context paragraphs of sufficient length, (4) propose four diverse and non--redundant policy options, and (5) annotate each policy with its primary capability restoration. To avoid extreme situations, we explicitly instructed the model not to invent unlikely events (e.g., large scale violence in low--crime areas) and to ``think like a local expert who has studied homelessness for many years''. Generation was performed separately for each location and scenarios were manually reviewed by our team of researchers and multi-disciplinary domain experts to ensure realism and eliminate incoherent answers.

\paragraph{Quality control.}
Generated scenarios were post-processed to remove redundancies and 
inaccuracies. When the LLM produced incomplete or duplicate content, we regenerated the scenario with a revised prompt. 
We also verified that the distribution of capabilities across policies was balanced, and that the language was neutral and devoid of harmful stereotypes.

\subsubsection{Policy Annotation and Ethical Grounding}
\label{subsec:policy-annotation}

Beyond the narrative context, every policy option in the benchmark is linked to one or more human capabilities. These tags 
serve two roles: (\emph{i}) they can guide evaluators (human or model) toward understanding the ethical aspirations behind each intervention, and (\emph{ii}) they provide a means for measuring whether LLM recommendations favor certain types of capabilities over others. Annotating the policies in the framework of the Capability Approach involved identifying the core goal of each policy --whether it primarily protects life and physical health, fosters social belonging, supports practical agency, or promotes other types of individual freedom -- and mapping that to the relevant capability. 

\subsubsection{Expert Annotation}
\label{subsec:human-eval}

To create a human
baseline against which LLMs can be compared, we engaged two domain experts (one primary expert and a secondary one) from three out of the four cities in the benchmark -- Barcelona, Johannesburg, and South Bend. 
Experts are multi-disciplinary scholars working on homelessness alleviation in 
non-profit organizations or in close collaboration with local non-profits in the cities in scope.

We provided experts with 10 out of the 40 scenarios relevant to their location as well as the 10 universal scenarios, asking them to prioritize the four policies in each scenario from 1 (most preferred) to 4 (least preferred). Experts were instructed that there was no single correct answer and that they should rank options based on their topic and local knowledge, as well as their ethical considerations. Through this process, we obtained six sets of human-derived rankings -- two for each city -- that serve as the benchmark’s human reference recommendations.

\subsubsection{Model Evaluation Tasks}
\label{subsec:llm-eval}

While the benchmark could be used for various analyses, here we are interested in studying and evaluating how humans or machines make decisions as policymakers in two primary modes. In the \emph{top--choice task}, they must select the single policy they would recommend; in the \emph{ranking task}, they are asked to produce an ordered ranking of all four policy options. 
We undertake both tasks using several LLMs spanning different parameter scales, as described next.



\subsection {LLMs as Policymakers}

Our homelessness mitigation benchmark is presented to LLMs, which are prompted to respond as decision makers representing a non-profit organization.
Here we describe our choice of LLMs and baseline prompts.

\paragraph{Models.}

We experiment with the following LLMs, listed in increasing order of number of model parameters: 
Granite 3.1 8B Instruct (8 B), GPT 4.1 Mini (8 B), Mixtral 8X7B Instruct ($\sim$ 47 B), Llama 3.3 70 B Instruct (70 B), Deepseek V3 (671 B), GPT 4.1 (estimated $\sim$ 1.8 T). These models capture a broad spectrum of characteristics, spanning a range of sizes and architectures, enabling richer comparisons.


\paragraph{Prompts.}

We prompt LLMs in two primary ways, one for each task. In one approach, we ask the LLM to select one policy among the $4$ choices per scenario;
in the second, we ask the LLM to rank their choices. In both cases, we also request the LLM to provide a brief justification for their response.
For the GPT models, we employ a constrained reasoning prompt that guides models through four steps: summarizing the dilemma, identifying capabilities restored by each policy, assessing pros and cons, and outputting a ranking with justification. 
In addition, we conduct some experiments with ablation studies where LLMs are asked to make choices using additional considerations, such as paying heed to contextual information. We refer the reader to Appendix C where we provide example prompts and prompt templates.

\subsection{Assessing Social Impact with ABMs}

\label{sec:ABM_pipeline}

We evaluate the social impact of policies chosen by LLMs vs. those proposed by human domain experts in the city of Barcelona, using an existing agent-model simulation which is purposely built to simulate homelessness policymaking in this city \cite{Aguilera2024}.



\paragraph{Architecture.}


We present a pipeline that uses LLMs as structured policy recommenders.
In this pipeline, LLMs convert the natural language policy proposals they have previously selected into agents' behavioral adjustments in the ABM. This is done by requesting the LLM to update the state-action-transition (SAT) matrix accordingly, which regulates the behavior of agents during the simulation.  
Due to space limitations, we relegate details about the mathematical foundations of the ABM pipeline to Appendix B.  This approach enables us to compare the effects of policies generated by LLMs and by domain experts in terms of needs satisfaction 
of the agents representing people experiencing homelessness (PEH).

    


\paragraph{Prompts.}
We prompt an LLM by giving a specific role of a public policy expert and providing the following information: a description of the SAT matrix structure, the policy to be evaluated, a step-by-step description on how the matrix can be technically modified and the indication that the policies are framed in the Capability Approach. The LLM produces an update of the SAT matrix, modifying the behavior of agents in the simulation. Prompt details are in Appendix C.

\section{Empirical Investigation}




\begin{figure*}[htb]
\setlength{\tabcolsep}{-0.05cm}
\begin{tabular}{ccc}
\includegraphics[width=.33\textwidth]{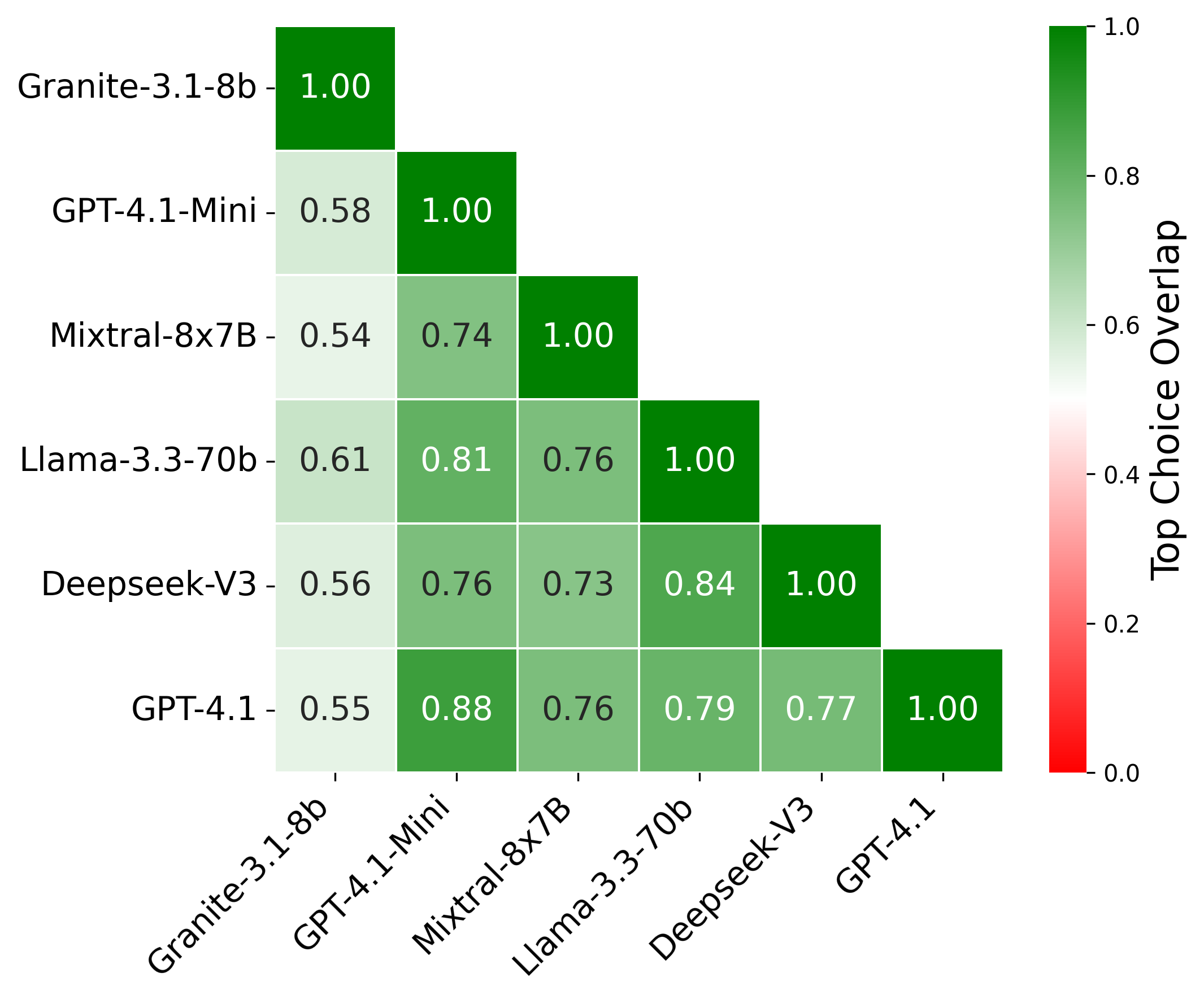}
&
\includegraphics[width=.33\textwidth]{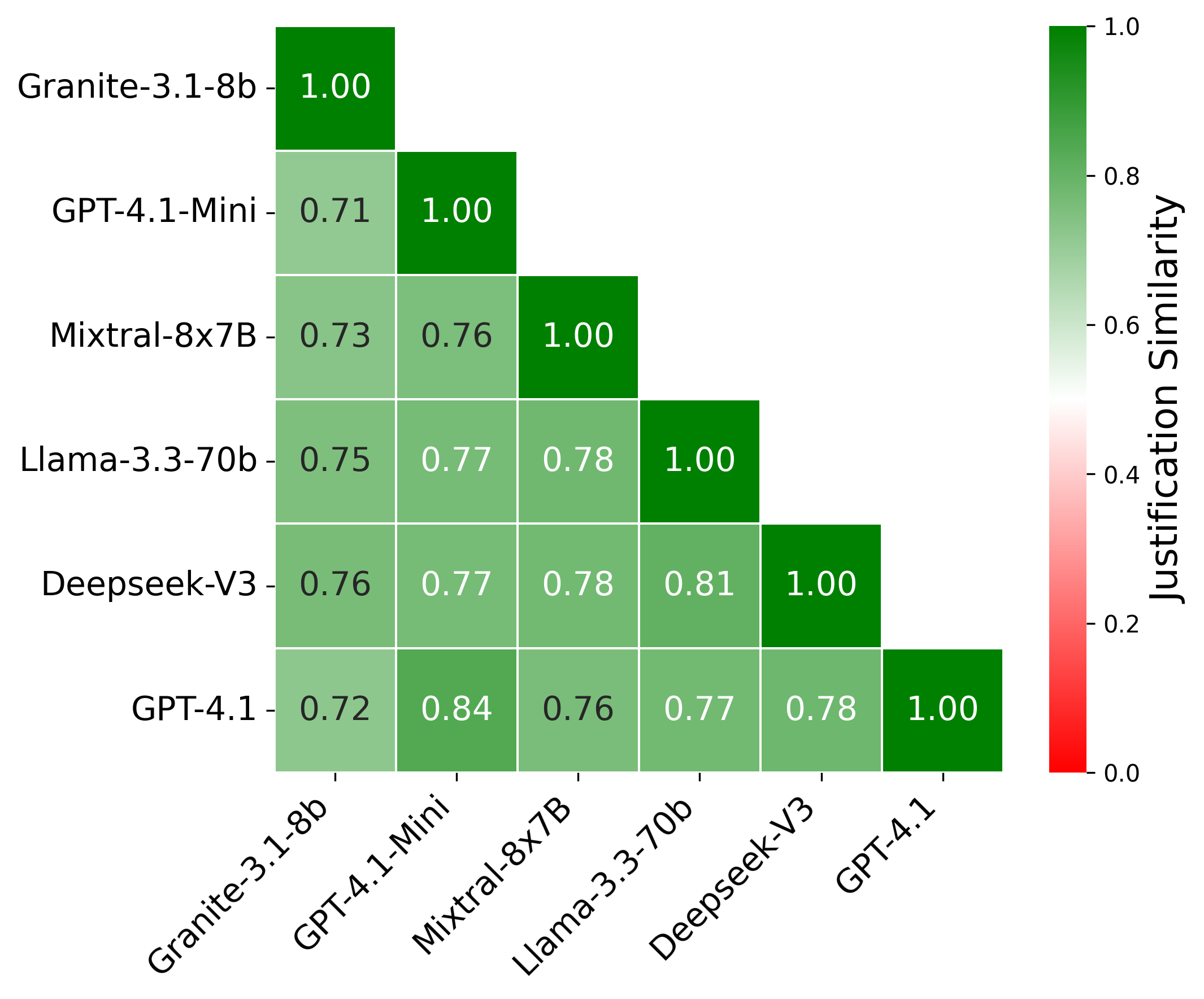}
&\includegraphics[width=.33\textwidth]{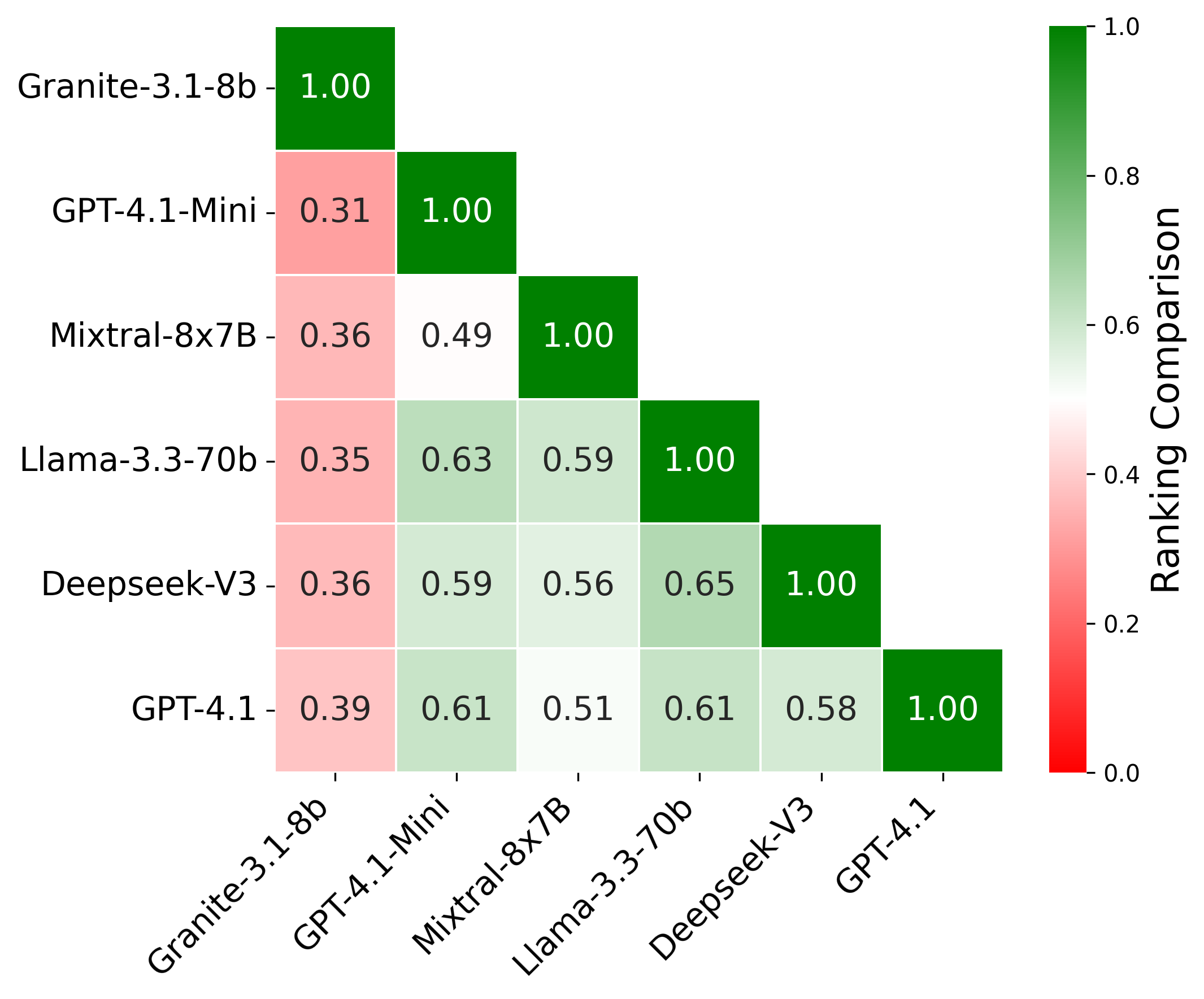}\\
{\footnotesize(a) Top Choice Overlap} & {\footnotesize(b) Justification Similarity}& {\footnotesize(c) Ranking Comparison}
\end{tabular}
\caption{Pairwise comparison of the choices and 
judgments of various LLMs using heat maps of the following: 
a) the fraction of common top choices, b) similarity between justifications for top choices (as measured using Sentence-BERT embeddings), and c) comparison of rankings of choices (as measured by the normalized Kendall tau distance).} 
\label{fig:comparing_LLMs} 
\end{figure*}

\subsection{Investigating LLMs' Choices}

\paragraph{Experiment.}
We explore how the various LLMs compare in terms of their choices on the homelessness mitigation benchmark.
Each LLM is asked to consider all $160$ contextualized decision scenarios ($40$ each across $4$ locations) and make a judgment about their top choice and rankings using neutral prompts.
It is important to note that there is no circularity in this evaluation -- while LLMs assisted in generating policy options for each scenario, the ranking task is fundamentally different: models must reveal their preferences among a fixed set of policies, thereby indirectly revealing how they prioritize different capabilities within the Capability Approach framework.
Figure~\ref{fig:comparing_LLMs} shows pairwise comparison plots between all $6$ LLMs under consideration in the form of heat maps, using different aggregate metrics across panels.
Panel (a) displays the fraction of scenarios where the top choices for a pair of models are identical. 
Panel (b)  shows the average semantic similarity between the textual justifications of LLMs for their top choice, as measured using Sentence-BERT embeddings.
Panel (c) compares rankings of choices, as gauged by the normalized Kendall tau distance, which is $1$ for identical rankings and $-1$ for completely reversed rankings.

\paragraph{Results.}
The plots in Figure~\ref{fig:comparing_LLMs} indicate that there may be disagreements between LLMs,  
particularly those from different model families. 
Panel (a) highlights that the smaller Granite 3.1 8B has less overlapping top choices, while other models choose the same policy around $70-80\%$ of the time.
Panel(b) suggests that semantic similarity between justifications is notably high across all pairs (0.71-0.84), indicating convergence in reasoning patterns and linguistic styles.
However, normalized Kendall tau correlations between full rankings are considerably lower (0.31-0.65) in panel (c), suggesting models use similar justification frameworks but apply different weightings when ordering options. 
These findings reveal that models agree most on how they justify, less on what they choose first, and least on how they rank the policies -- perhaps because surface-level homogenization in articulation masks diverse underlying preferences.

\subsection{Comparing Choices of LLMs vs. Experts}

\paragraph{Experiment.}

Our comparison with experts operates on two levels: agreement on the top-ranked policy choice 
as well as the
ranking over four policies, measured using the normalized Kendall tau distance. 
The comparison between the 
human experts on Universal scenarios provides a reference for inter-human agreement on decontextualized policy dilemmas. 


\begin{figure*}[htb]
\centering
\includegraphics[width=\linewidth]{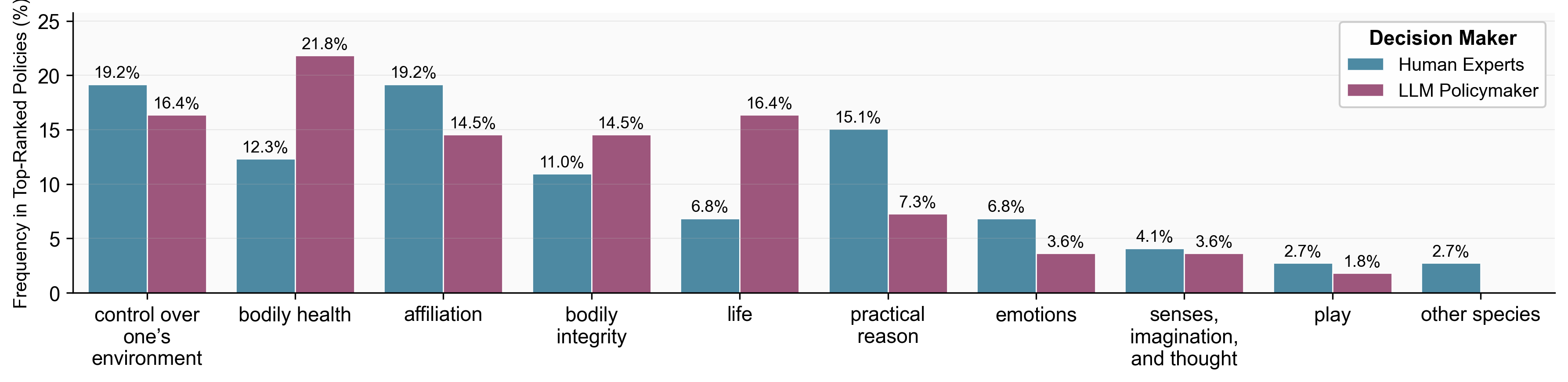}
\caption{Comparison of capabilities prioritized in the top policy choices of human experts and GPT-4.1 across all scenarios.}
\label{fig:capability_comparison}
\end{figure*}

\begin{table}[h!]
\centering
\caption{Alignment Scores: GPT-4.1 vs. Experts. Location-specific results represent the average alignment across multiple domain experts. Inter-Expert Consensus rows indicate the agreement between the human experts themselves.}
\label{tab:alignment_scores_compact}
\small 
\begin{tabular*}{\columnwidth}{@{\extracolsep{\fill}} l c c}
\toprule
\textbf{Comparison Pair} & \textbf{Top Choice} & \textbf{Ranking} \\
& \textbf{Agreement} & \textbf{Corr. ($\tau$)} \\
\midrule
\multicolumn{3}{l}{\textit{Location-Specific Scenarios}} \\
\quad GPT-4.1 vs. BCN Experts (Avg.) & 50\% & 0.32 \\
\quad \quad \textit{BCN Inter-Expert Consensus} & 50\% & 0.23 \\
\quad GPT-4.1 vs. JHB Experts (Avg.) & 55\% & 0.28 \\
\quad \quad \textit{JHB Inter-Expert Consensus} & 50\% & 0.50 \\
\quad GPT-4.1 vs. SB Experts (Avg.) & 50\% & 0.22 \\
\quad \quad \textit{SB Inter-Expert Consensus} & 20\% & 0.17 \\
\midrule
\multicolumn{3}{l}{\textit{Universal Scenarios}} \\
\quad GPT-4.1 vs. Exp. 1 (JHB) & 20\% & 0.13 \\
\quad GPT-4.1 vs. Exp. 2 (SB) & 60\% & 0.33 \\
\quad GPT-4.1 vs. Exp. 3 (BCN) & 40\% & -- \\
\quad \textbf{Human Inter-Expert Consensus} & \textbf{33\%} & \textbf{0.07} \\
\bottomrule
\end{tabular*}
\end{table}

\paragraph{Results.}

First, we compare GPT-4.1 with multiple experts across urban contexts. As shown in Table~\ref{tab:alignment_scores_compact}, alignment varies by city but frequently meets or exceeds inter-expert consensus. In Johannesburg, the LLM’s top choice matched expert priorities in 55\% of cases ($\tau = 0.28$), exceeding the 50\% consensus between experts. In South Bend, the model’s 50\% top-choice agreement ($\tau = 0.22$) also outperformed the 20\% consensus between experts. Barcelona showed 50\% alignment ($\tau = 0.32$), matching the consensus between experts.


The Universal scenarios provides a critical reference point for these findings. Comparing three experts across regions shows a low average consensus of 33\% ($\tau = 0.07$), reflecting the importance of local context in policy prioritization. In this light, GPT-4.1’s strong alignment with the South Bend expert on these universal dilemmas (60\% agreement, $\tau = 0.33$) is notable. This disparity could potentially be illustrating a bias, possibly stemming from the over-representation of US-centric published discourse in their training data. Conversely, the lower agreement with experts from Johannesburg (20\%) and Barcelona (40\%) on identical universal dilemmas may indicate
a gap in the model's ability to align with the local contexts, regarding the topic of homelessness policymaking, in geographic locations from where there is less training data in frontier models. 
However, this pattern should not be over-interpreted. 
Differences between experts' opinions may arise due to heterogeneity in their underlying policy philosophies, or due to the way Universal scenarios are contextualized. We therefore treat this as a hypothesis that motivates further investigation rather than a 
claim about the nature of the training-data composition.

\begin{figure}[tb]
\centering
\setlength{\tabcolsep}{-0.05cm}  
\begin{tabular}{c}
\includegraphics[width=0.45\textwidth]{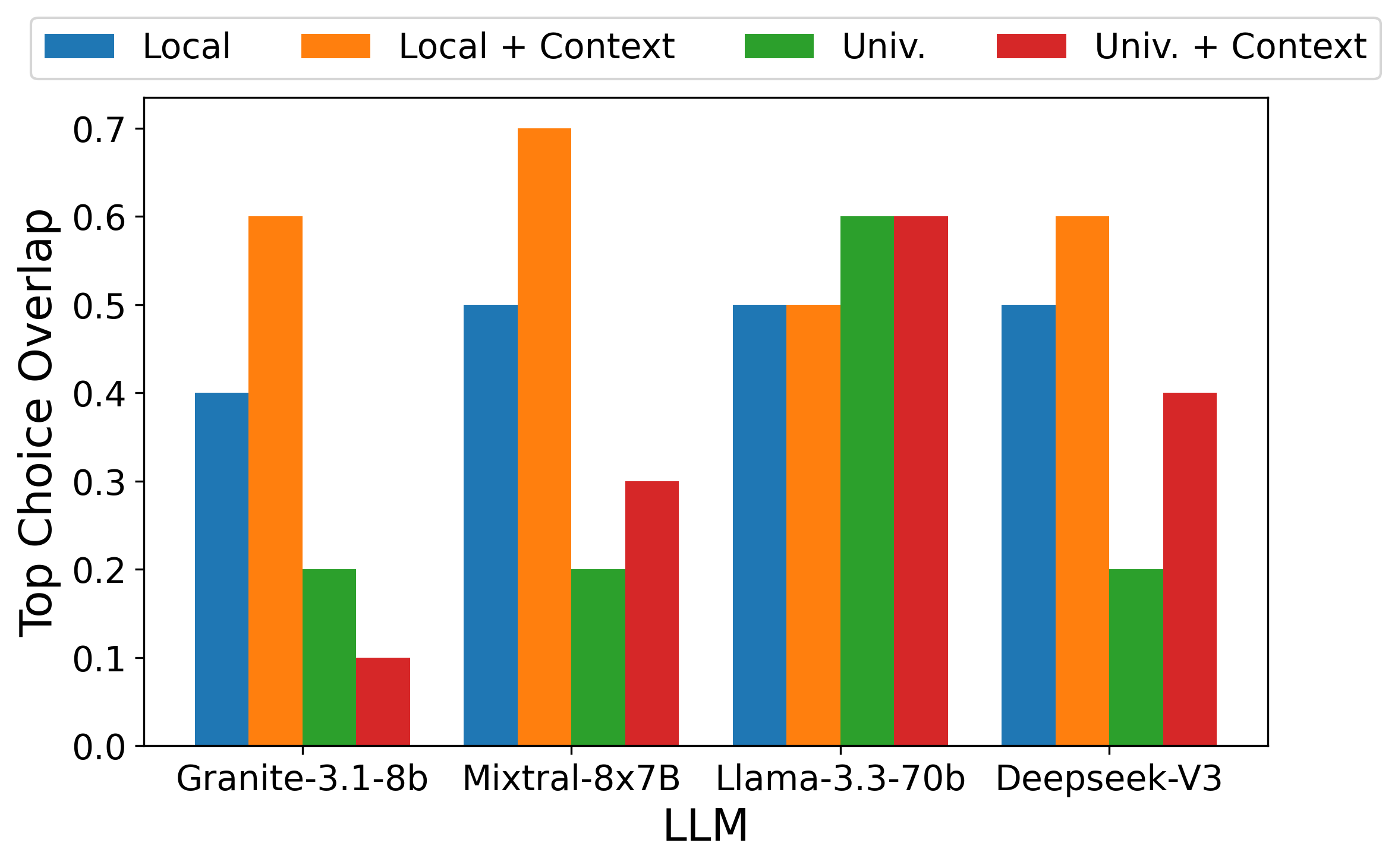}
\end{tabular}
\caption{Comparing top choice overlap between 4 LLMs and the primary domain expert in Johannesburg, with and without prompting LLMs to consider the local context when selecting policies.}
\label{fig:context}
\end{figure}

\begin{table*}[h]
\centering
\caption{Comparison of of how policies recommended by LLMs and experts fulfill the needs of PEH agents in the simulation.}
\label{tab:combined_changes}
\adjustbox{max width=\textwidth}{
\begin{tabular}{l*{4}{ccc}}
\toprule
\multirow{2}{*}{Scenario} & \multicolumn{3}{c}{Physiological} & \multicolumn{3}{c}{Safety} & \multicolumn{3}{c}{Belonging} & \multicolumn{3}{c}{Self-esteem} \\
\cmidrule(lr){2-4} \cmidrule(lr){5-7} \cmidrule(lr){8-10} \cmidrule(lr){11-13}
 & Mean & Std. & p-value & Mean & Std. & p-value & Mean & Std. & p-value & Mean & Std. & p-value \\
\midrule
\textbf{Scenario 1} \\
\quad LLM Policy & +0.019 & -0.005 & 0.001 & +0.032 & -0.023 & 0.001 & +0.021 & -0.008 & 0.030 & +0.037 & -0.025 & 0.001 \\
\quad Expert Policy & -0.004 & -0.004 & 0.457 & +0.026 & -0.023 & 0.006 & -0.036 & -0.011 & 0.009 & +0.027 & -0.025 & 0.015 \\
\addlinespace
\textbf{Scenario 3} \\
\quad LLM Policy & +0.012 & -0.010 & 0.080 & +0.031 & -0.025 & 0.001 & +0.012 & -0.013 & 0.302 & +0.036 & -0.027 & 0.001 \\
\quad Expert Policy & +0.011 & -0.011 & 0.029 & +0.030 & -0.025 & 0.001 & +0.009 & -0.017 & 0.092 & +0.035 & -0.028 & 0.001 \\
\addlinespace
\textbf{Scenario 5} \\
\quad LLM Policy & +0.014 & -0.020 & 0.134 & +0.031 & -0.026 & 0.002 & +0.014 & -0.034 & 0.322 & +0.038 & -0.030 & 0.002 \\
\quad Expert Policy & +0.002 & -0.020 & 0.909 & +0.029 & -0.025 & 0.002 & -0.016 & -0.032 & 0.036 & +0.033 & -0.029 & 0.003 \\
\bottomrule
\end{tabular}
}
\end{table*}


We further investigate the importance of contextualization in Figure~\ref{fig:context}, where we consider the scenarios in Johannesburg, comparing top choices of the four non-GPT LLMs with those from the primary expert in four ways: over the 10 contextualized (Local) scenarios and the 10 Universal scenarios, both with and without additional prompting to the LLM to explicitly consider the local geographical context. 
We observe that: 1) there is higher alignment with the domain expert choice in the Local rather than Universal settings, and 2) specifying the context impacts LLMs' choices and yields higher alignment, even when scenarios are Local, i.e., already somewhat contextualized by design. 

Beyond policy choice agreement rates, we find systematic divergence in the \textit{types} of capabilities prioritized, as visualized in Figure~\ref{fig:capability_comparison}. Experts' top choices frequently centered on policies restoring \textbf{practical reason} (e.g., skills training, participatory governance) and \textbf{affiliation} (e.g., peer support, community integration). In contrast, GPT-4.1 demonstrated a stronger tendency to select policies that protect against immediate physical threats, prioritizing \textbf{life}, \textbf{bodily health}, and \textbf{bodily integrity}. For instance, in the scenario titled ``Safe-Parking for Vehicle Dwellers'' (South Bend), the human expert preferred `Conversion-to-Housing Incentive Grants` to transition residents into stable rooms, emphasizing long-term capability expansion. GPT-4.1 favored a `Monitored Safe-Parking Lot`, echoing its preference toward scalable, immediate 
hazard reduction. Further studies should be conducted to identify if LLMs' preferences pertaining to the prioritization of safety-first interventions over relational or structural ones is connected to the over-representation of data from specific parts of the world and the public narrative on homelessness in these regions. 

\subsection{Gauging the Impact of Policies using ABMs}
\label{subsec:impactofpolicies}

\paragraph{Experiment.}

We evaluate benchmarked policies by measuring needs satisfaction for agents representing persons experiencing homelessness (PEH) within our ABM pipeline. Simulations use 80 agents over 1,450 steps (simulating two months). We employ Deepseek-R1 ($T=0.1$) to update the SAT matrix for Barcelona-specific scenarios where expert and GPT-4.1 policy prioritizations diverge (Scenarios 1, 3, and 5). While this framework is currently tailored to Barcelona, future work will include three additional cities (see Appendix B). We compare 10 baseline simulations using the original SAT matrix against 10 simulations using the LLM-updated matrix. This approach highlights how expert-selected versus LLM-suggested policies differ in addressing PEH needs satisfaction.

\paragraph{Results.}

Table \ref{tab:combined_changes} summarizes the changes observed in the ABM simulation when applying either the policies recommended by LLMs or by the domain experts. For each category of needs, `Mean' represents the difference of the average satisfaction value for PEH agents with or without the policy. `Std' represents the difference in the dispersion of needs satisfaction with or without policies. And finally, `p-value' is associated with a standard t-test to assess whether the changes in terms of needs' satisfaction were statistically significant.

Based on results in Table \ref{tab:combined_changes}, LLM-recommended policies seem to demonstrate a slightly superior overall balance compared to expert-recommended ones. While both approaches notably enhance safety and self-esteem needs, improvements from LLM choices appear to be more substantial. 
Both policy types successfully reduced outcome variability (negative change in std values), but LLMs achieved this while maintaining positive or neutral effects across all need categories. More detailed results are available in Appendix D.







\section{Discussion and Conclusion}

This paper investigates the potential of LLMs to support policymaking on the subject of homelessness alleviation in four global cities, using a benchmark grounded in the Capability Approach for human development and supported by an agent-based simulation. The study is motivated by the expressed need of government officials, domain experts and non-profit organizations, who need to make informed decisions in a complex and evolving environment, often with limited time and data. The capacity of LLMs to leverage vast amounts of humanly-generated existing data and explore flexible role-play scenarios (i.e. acting as domain experts) make them uniquely suited to potentially provide new insights for the complexity of social policymaking. However, given the current limitations of LLMs, in terms of transparency, explainability, security and ethics, among others, there are open questions around whether policies recommended by LLMs are to be trusted and can be helpful for decision makers. 


Our empirical investigation about LLM policy recommendations suggests that while models may have learned similar ways to articulate reasoning, they maintain diversity in how they weight and order the options. 
Importantly, LLMs' choices seem to prioritize immediate physical safety across the different contexts, over policies that aim to improve the social affiliation of the affected population and mitigate ostracism. This could be exposing a context-blind rigidity and a potential bias based on the published online discourse on homelessness in the regions of the world with over-represented data in LLMs. Experts, in contrast, demonstrably tailored their decisions to address locally encoded sociopolitical realities that can aggravate the ostracism faced by PEHs, such as related to ethnicity (in South Africa) or to religious minorities (in South Bend). 
However, for a selection of the scenarios in the benchmark simulated in the ABM, our findings show that the social impact of policies selected by LLMs  slightly better fulfill the overall needs of PEH population, vs. those selected by the domain expert in Barcelona. 



As is the case in most computational approaches to complex social topics, our work has several limitations that need to be addressed in future work.  First of all, a wider diversity of non-profit organizations and domain experts should be included in future policy scenario creation and prioritization.  
Even though the baseline scenario was created in close collaboration with domain experts, the remaining scenarios in the benchmark rely heavily on LLM assistance, albeit with detailed designed prompting that incorporates the Capability Approach and the local contexts. Furthermore, benchmark scenarios and policies are currently all in English.
We recommend that future studies build the entire benchmark (not only the baseline) with realistic policies proposed by domain experts in the local languages of the cities in scope. 
Finally, the agent-based modeling framework that allows one to evaluate the impact of the policies suggested by LLMs, versus those recommended by domain experts, currently handles the policy scenarios for one of the locations in scope only (Barcelona). The pipeline in between LLM generated policies and the ABM framework aims to open up a line of work to evaluate LLM-suggested policies through ABM frameworks in the different locations in scope at scale.  

Our current findings highlight both promise and caution for the use of LLMs to inform social policymaking. On the one hand, LLMs show a certain degree of alignment with domain experts and propose plausible policy recommendations, which can be valuable inputs to policy makers and non-profit organizations in the context of limited data and time constraints. On the other, our findings illustrate that LLMs are less sensitive to local contexts than domain experts, which could be influenced by the over-representation of online data from certain parts of the world. In addition, LLMs seem to not sufficiently consider the restoration of central human development capabilities (such as the need for affiliation) as compared to domain experts, who emphasized measures dedicated to mitigate social ostracism against PEHs. 
Moving forward, the responsible use of of LLMs in social policymaking contexts will require not only  technical refinement, but deliberate mechanisms of evaluation in line with conceptual ethical frameworks, such as the Capability Approach of human development, as well as responsiveness to local contexts and valuable inputs from domain experts. 

\clearpage

\bibliographystyle{named}
\bibliography{references}

\appendix
\onecolumn

\begin{appendices}

\section{Nussbaum's Central Human Capabilities}
\label{sec:app-capabilities}

We list and describe the 10 central human capabilities in the Capability Approach, based on \cite{nussbaum2011}.

\begin{itemize}
    \item \textbf{Life} : Being able to live to the end of a human life of normal length; not dying prematurely, or before one's life is so reduced as to be not worth living.
    \item \textbf{Bodily Health} : Being able to have good health, including reproductive health; to be adequately nourished; to have adequate shelter.
    \item \textbf{Bodily Integrity} : Being able to move freely from place to place; to be secure against violent assault, including sexual assault and domestic violence; having opportunities for sexual satisfaction and for choice in matters of reproduction.
    \item \textbf{Senses, Imagination and Thought} : Being able to use the senses, to imagine, think, and reason—and to do these things in a "truly human" way, a way informed and cultivated by an adequate education, including, but by no means limited to, literacy and basic mathematical and scientific training. Being able to use imagination and thought in connection with experiencing and producing works and events of one's own choice, religious, literary, musical, and so forth. Being able to use one's mind in ways protected by guarantees of freedom of expression with respect to both political and artistic speech, and freedom of religious exercise. Being able to have pleasurable experiences and to avoid non-beneficial pain.
    \item \textbf{Emotions} : Being able to have attachments to things and people outside ourselves; to love those who love and care for us, to grieve at their absence; in general, to love, to grieve, to experience longing, gratitude, and justified anger. Not having one's emotional development blighted by fear and anxiety. (Supporting this capability means supporting forms of human association that can be shown to be crucial in their development.)
    \item \textbf{Practical Reason} :  Being able to form a conception of the good and to engage in critical reflection about the planning of one's life. (This entails protection for the liberty of conscience and religious observance.)
    \item \textbf{Affiliation} : Being able to live with and toward others, to recognize and show concern for other humans, to engage in various forms of social interaction; to be able to imagine the situation of another. (Protecting this capability means protecting institutions that constitute and nourish such forms of affiliation, and also protecting the freedom of assembly and political speech.)
    Having the social bases of self-respect and non-humiliation; being able to be treated as a dignified being whose worth is equal to that of others. This entails provisions of non-discrimination on the basis of race, sex, sexual orientation, ethnicity, caste, religion, national origin and species.
    \item \textbf{Other Species} : Being able to live with concern for and in relation to animals, plants, and the world of nature.
    \item \textbf{Play} : Being able to laugh, to play, to enjoy recreational activities.
    \item \textbf{Control Over One's Environment} : 
    Being able to participate effectively in political choices that govern one's life; having the right of political participation, protections of free speech and association.
    Material. Being able to hold property (both land and movable goods), and having property rights on an equal basis with others; having the right to seek employment on an equal basis with others; having the freedom from unwarranted search and seizure. In work, being able to work as a human, exercising practical reason and entering into meaningful relationships of mutual recognition with other workers.
\end{itemize}

\section{Urban ABM Backbone: Theoretical Foundations}
\label{Maths_foundations}

This ABM framework comes from the work of \cite{Aguilera2024}. It relies on the set of Maslow's need categories (
Physiological, Safety, Belonging, Esteem) as $C$ and
the set of needs within each category $c \in C$ as $N_c$ (Physiological: food, shelter, sleep, health; 
Safety: clothing, financial security, employment, education; 
Belonging: family, friendship, intimacy; Esteem: freedom, status, self-esteem) thus resulting in 14 needs across each category.
Each agent $a_i \in \mathcal{A}$ maintains:
\begin{itemize}
    \item A status $s \in \{\texttt{homeless}, \texttt{employed}, \texttt{unemployed}\}$
    \item A need-satisfaction vector $\mathbf{n}^{(t)} \in [0,1]^{14}$ at timestep $t$
    \item A satisfaction-Action Transition (SAT) matrix based on each status $\mathbf{M}_a \in \mathbb{R}^{14 \times 11}$. This SAT matrix maps the 11 possible actions (See Appendix) that agents can take depending on status to a need satisfaction bonus. 
\end{itemize}
An importance function maps every need category to the weight that the agent assigns to it, $\operatorname{Imp} : C \to [0, 1]$. At time-step $t$, the need satisfaction
level of need $n \in N_c$ (for some $c \in C$) is given by the need satisfaction level function $\operatorname{NSL}_t : 
\bigcup_{c \in C} N_c \to [0, 1]$, which maps every need (across all categories) to its current degree of fulfilment. It can
be written as an iterative function:

\begin{align*}
\operatorname{NSL}_t(n) &= \gamma_{n,s(n)} \cdot \operatorname{NSL}_{t-1}(n)
\end{align*}

where $\gamma_{n,s} \in [0, 1]$ is the decay rate for a need $n$ and an agent with
status $s$. The customization of the decay rates based on the nature of need and the
agent's status $s$ allows to create a differentiation between the personal circumstances of the agents.
\\
Agent decision-making then follows a greedy algorithm. At each timestep $\ t = 1$ hour, agent $a$ with status $s$:
\begin{enumerate}
    \item Identifies dominant unmet need: $n^* = \operatorname*{arg\,min}_j \, \mathbf{n}_j^{(t)}$
    \item Selects action $\alpha$ via:
    \begin{equation}
\alpha_t = \underset{\alpha \in A}{\mathrm{arg}\,\mathrm{max}} 
\left[ 
    \sum_{c \in C}
    \left(
        \sum_{\mathclap{n \in N_c}} M_{a}[n] \cdot (1-NSL_{t}(n)
    \right)
    \cdot Imp(c)
\right].
\tag{3}
\end{equation}
    
    \item Updates needs: \\ $\mathbf{n}_j^{(t+1)} = \gamma_{{n}_j,s} \mathbf{n}_j^{(t)} + \delta_{j,n^*} \mathbf{M}_a[n^*, \alpha]$
    \\
    with $\delta_{j,n^*}$ the Kronecker delta symbol :
\[
\delta_{j,n^*} = 
\begin{cases} 
1 & \text{if } j = n^*  \\
0 & \text{if } j \neq n^* 
\end{cases}
\]
\end{enumerate}
Thus the SAT Matrix is the key that is used by the agents to inform their behaviors by selecting the action to accomplish at each step of the simulation.
\subsubsection*{LLM-ABM Integration}

The critical innovation that we bring to address the problem of scalability for policy testing lies in bridging \textit{social policy narratives} with \textit{expected behavioral adjustments} through LLM translation.
\\
Indeed usual policy testing using ABM through SAT matrices faces two fundamental limitations:

\begin{equation}
\underbrace{\mathcal{P}_{\textrm{narrative}}}_{\substack{\text{Natural language} \\ \text{policy proposals}}} \xrightarrow{\textit{manual}} \underbrace{\Delta \mathbf{M}}_{\substack{\text{SAT matrix} \\ \text{adjustments}}} \quad \text{is} \quad \begin{cases} 
\text{work intensive} \\
\text{opaque in value alignment}
\end{cases}
\end{equation}

While informing the behavior of agents by calling LLMs for each decision of each agent during a simulation can be a technical solution instead of relying on a SAT matrix, this approach appears to be constrained by scalability concerns and limited number of tokens context for LLMs. Indeed, numerous LLMs call to simulate the interaction between agents result in a large amount of tokens to be processed and generated by LLMs, which in returns make the simulation costly in terms of money (if you need an access to a private API), energy consumption, and time \cite{feng-etal-2025-agentmove}.

Our pipeline addresses this by formalizing LLMs as \textit{differential policy operators}:
\begin{align}
\mathcal{P}_{\textrm{narrative}} \xrightarrow{\textit{LLM}_{\theta}} \Delta \mathbf{M} \quad \textrm{s.t.} \quad & \|\Delta \mathbf{M}\|_\infty \leq 0.03 \nonumber 
\label{eq:constraints}
\end{align}
where $\theta$ denotes ethical priors (capability approach) encoded in the LLM's prompt (See appendix).

\subsubsection*{Causal Interpretability Framework}
The integration establishes a closed-loop causal pathway for ethical impact assessment:
\begin{equation}
\underbrace{\mathcal{P}}_{\text{Policy}} \xrightarrow{\textit{LLM}} \underbrace{\Delta \mathbf{M}}_{\text{Adjustment}} \xrightarrow{\textit{ABM}} \underbrace{\Delta \mathbb{E}[\alpha]}_{\text{Behavior}} \xrightarrow{} \underbrace{\Delta \mathbb{E}\left[\mathbf{n}^{(T)} \mid s=\texttt{homeless}\right]}_{\text{Equity Impact}} \label{eq:causal_chain}
\end{equation}

Policies $\mathcal{P}$ thus inject perturbations via precondition (e.g., $s = \texttt{homeless}$) and SAT overrides:
\begin{equation}
\mathbf{M}_{a} \leftarrow 
\begin{cases} 
\mathbf{M}_{a} + \Delta \mathbf{M} & \text{if } \phi(a) = \texttt{True} \\
\mathbf{M}_{a} & \text{otherwise}
\end{cases}
\end{equation}
enabling targeted assessment of distributional effects while preserving mechanistic accountability through the ABM's core. It shows that the policies can used to target other groups of people than just people experiencing homelessness, thus expanding the potential usage of this framework.

\clearpage

\section{Prompts for LLMs}
\label{sec:app-prompts}

\subsection{Prompts for LLM Responses for the Benchmark}

\paragraph{Baseline prompt for top choices.} Below, we provide an example prompt for requesting the LLM to choose from among the policies in a decision scenario in the homelessness mitigation benchmark. This example is for the first decision scenario in Barcelona from the benchmark.

\rule{\linewidth}{0.4pt}\\
\lstset{
    basicstyle=\ttfamily\small,
    escapeinside={(*@}{@*)},
    breaklines=true,
    backgroundcolor=\color{gray!10},
}
\begin{lstlisting}
Task: 
Act as an expert policymaker. Your goal is to study the decision scenario and to propose your top policy among the policy choices. 

Instructions: 
1. Decision Scenario Analysis: 
   - Analyze the context of the decision scenario as well as the potential policy choices for helping alleviate homelessness. 

2. Response Structure: 
   - Answer: Start with the policy choice number wrapped in squared brackets (e.g., [1], [2], [3], [4]) based on your analysis. 
   - Justification: Briefly explain why you chose the selected option.

**Input Format:** 

Scenario: 
{ Insert the scenario name and context here }

Policy Options: 

 {Insert the list of policy options along with their descriptions }

--- 
Your Turn: 

Scenario: 
Title: Deploying EU Recovery Funds: Temporary Modular Housing vs. Long-Term Solutions
Context: Barcelona has received (*@\texteuro@*)24 million in NextGenerationEU recovery funds earmarked for "urgent housing innovation." City planners propose using part of the money to expand APROP(*@\textemdash@*)Barcelona(*@\textquoteright@*)s experimental shipping-container micro-apartment program(*@\textemdash@*)so that unsheltered residents near Pla(*@\c{c}@*)a de les Gl(*@\`{o}@*)ries can move into 92 prefabricated units within eighteen months. Critics argue the funds should instead reinforce permanent Housing First placements or bolster rent-supplement vouchers in adjacent municipalities where rents are lower. Local neighborhood groups worry that another modular block will accelerate gentrification pressures and erase remaining community gardens created on the former elevated ring-road site. With municipal elections a year away, the non-profit (*@\textquotedblright@*)Habitat Digne BCN(*@\textquotedblright@*) must pick one strategic path to recommend before City Hall freezes the allocation. 

Policy Options: 
1: Scale Up APROP to 92 Additional Units at Gl(*@\`{o}@*)ries
Dedicate the full (*@\texteuro@*)24 million to fast-track construction of stackable, energy-efficient container micro-homes on municipally owned land beside the new tramway link. Each 30 m(*@\textsuperscript{2}@*) unit includes private bathrooms, kitchenette, and climate control. A social-work team would provide on-site case management for up to five years, prioritizing rough sleepers with chronic illnesses, thereby stabilizing residents(*@\textquoteright@*) health and safety while the district(*@\textquoteright@*)s formal social-housing queue progresses. 
2: Bolster Permanent Housing First Placements across AMB
Redirect two-thirds of the grant toward purchasing existing flats in low-vacancy municipalities like Santa Coloma and Sant Adri(*@\`{a}@*), integrating tenants directly into mainstream neighborhoods with individualized support. The remaining funds seed a revolving rent-guarantee pool that reassures private landlords, empowering participants to sign standard leases and exercise normal tenancy rights rather than residing in temporary structures. 
3: Rent-Supplement Vouchers Plus Job-Link Services
Create 600 time-limited housing vouchers pegged to 40 % of median rent, valid city-wide but portable across metropolitan transit zones. Pair vouchers with tailored employment coaching at Barcelona Activa hubs, aiming to lift recipients above the poverty threshold within two years. Although less capital-intensive, this path assumes the private market can absorb increased demand without driving rents.
4: Community Land Trust & Garden Preservation Compact
Invest the funds in purchasing the contested Gl(*@\`{o}@*)ries plots for a resident-led community land trust that mixes 50 permanent cooperative apartments with ground-floor urban farms and cultural studios. This hybrid model keeps green commons intact, promotes ecological stewardship of pollinator gardens, and embeds arts-based therapy workshops, supporting holistic well-being for formerly unhoused residents while resisting speculative real-estate cycles.
 

Respond using the structured format as described above.
\end{lstlisting}
\rule{\linewidth}{0.4pt}\\

\paragraph{Variation for ranking.}

In a variation, we request the LLM to provide a ranking among the choices in a decision scenario. The only change in the prompt lies in our primary instruction to the LLM for the task, which is as follows:
``Act as an expert policymaker. Your goal is to study the decision scenario and to propose an ordered ranking of the provided policy choices, from most preferred to least preferred. ''

\paragraph{Variation for contextual emphasis.}

In a variation, we request the LLMs to pay heed to the context as they make their choice. The only change in the prompt lies in our primary instruction to the LLM for the task, where we append some text. An example of the instruction (for Johannesburg) is as follows: ``Act as an expert policymaker. Your goal is to study the decision scenario and to propose your top policy among the policy choices. This decision scenario is set in the city of Johannesburg in South Africa. Please try to take the location into account for your analysis.''

\subsection{ABM Pipeline Prompt}
\label{sec:app-pipeline-prompt}

Here we share our prompt for using LLMs to connect policies to social impact using ABMs.

\rule{\linewidth}{0.4pt}\\
\lstset{
    basicstyle=\ttfamily\small,
    escapeinside={(*@}{@*)},
    breaklines=true,
    backgroundcolor=\color{gray!10},
}

\begin{lstlisting}
You are a public policy expert with deep knowledge of social justice, urban development,
and the CAPABILITY APPROACH developed by Martha Nussbaum.
You specialize in designing realistic and ethically sensitive agents-based MODELS for use in simulated environments involving large language models (LLMs).

Here you are given a SAT matrix mapping needs to actions for people experiencing homelessness.

MATRIX STRUCTURE EXPLANATION:
-The matrix has 14 rows (needs) and 11 columns (actions)
- matrix[row][column] = matrix[need(*@\textunderscore@*)index][action(*@\textunderscore@*)index] = satisfaction(*@\textunderscore@*)value
- Row 0 = (*@\textquotedblright@*)food(*@\textquotedblright@*), Row 1 = (*@\textquotedblright@*)shelter(*@\textquotedblright@*), Row 2 = (*@\textquotedblright@*)sleep(*@\textquotedblright@*), etc. 
- Column 0 = (*@\textquotedblright@*)go(*@\textunderscore@*)grocery(*@\textquotedblright@*), Column 9 = (*@\textquotedblright@*)go(*@\textunderscore@*)reception(*@\textunderscore@*)center(*@\textquotedblright@*), etc. 

(*@\textbraceleft@*)
  (*@\textquotedblright@*)actions(*@\textquotedblright@*): [(*@\textquotedblright@*)go(*@\textunderscore@*)grocery(*@\textquotedblright@*), (*@\textquotedblright@*)go(*@\textunderscore@*)hospital(*@\textquotedblright@*), (*@\textquotedblright@*)go(*@\textunderscore@*)shopping(*@\textquotedblright@*), (*@\textquotedblright@*)go(*@\textunderscore@*)leisure(*@\textquotedblright@*), (*@\textquotedblright@*)invest(*@\textunderscore@*)education(*@\textquotedblright@*), (*@\textquotedblright@*)sleep(*@\textunderscore@*)street(*@\textquotedblright@*), (*@\textquotedblright@*)beg(*@\textquotedblright@*), (*@\textquotedblright@*)steal(*@\textunderscore@*)food(*@\textquotedblright@*), (*@\textquotedblright@*)steal(*@\textunderscore@*)clothes(*@\textquotedblright@*), (*@\textquotedblright@*)go(*@\textunderscore@*)reception(*@\textunderscore@*)center(*@\textquotedblright@*), (*@\textquotedblright@*)go(*@\textunderscore@*)prison(*@\textquotedblright@*)],
  (*@\textquotedblright@*)needs(*@\textquotedblright@*): [(*@\textquotedblright@*)food(*@\textquotedblright@*), (*@\textquotedblright@*)shelter(*@\textquotedblright@*), (*@\textquotedblright@*)sleep(*@\textquotedblright@*), (*@\textquotedblright@*)health(*@\textquotedblright@*), (*@\textquotedblright@*)clothing(*@\textquotedblright@*), (*@\textquotedblright@*)financial security(*@\textquotedblright@*), (*@\textquotedblright@*)employment(*@\textquotedblright@*), (*@\textquotedblright@*)education(*@\textquotedblright@*), (*@\textquotedblright@*)family(*@\textquotedblright@*), (*@\textquotedblright@*)friendship(*@\textquotedblright@*), (*@\textquotedblright@*)intimacy(*@\textquotedblright@*), (*@\textquotedblright@*)freedom(*@\textquotedblright@*), (*@\textquotedblright@*)status(*@\textquotedblright@*), (*@\textquotedblright@*)self-esteem(*@\textquotedblright@*)],
  (*@\textquotedblright@*)matrix(*@\textquotedblright@*): [  
    [1.0, 0.0, 0.0, 0.4, 0.0, 0.0, 0.15, 0.7, 0.0, 0.5, 0.0],  
    [0.0, 0.0, 0.0, 0.0, 0.0, 0.0, 0.0, 0.0, 0.0, 0.7, 0.0],  
    [0.0, 0.0, 0.0, 0.0, 0.0, 0.7, 0.0, 0.0, 0.0, 0.0, 0.0],  
    [0.3, 1.0, 0.0, 0.0, 0.0, 0.0, 0.0, 0.0, 0.0, 0.0, 0.0],  
    [0.0, 0.0, 1.0, 0.0, 0.0, 0.0, 0.0, 0.0, 0.8, 0.0, 0.0],  
    [0.0, 0.0, 0.0, 0.0, 0.5, 0.0, 0.5, 0.0, 0.0, 0.0, 0.0],  
    [0.0, 0.0, 0.0, 0.0, 0.5, 0.0, 0.5, 0.0, 0.0, 0.0, 0.0],  
    [0.0, 0.0, 0.0, 0.0, 1.0, 0.0, 0.0, 0.0, 0.0, 0.0, 0.0],  
    [0.0, 0.0, 0.0, 0.0, 0.0, 0.0, 0.0, 0.0, 0.0, 0.0, 0.0],  
    [0.0, 0.0, 0.0, 0.4, 0.0, 0.0, 0.0, 0.0, 0.0, 0.0, 0.0],  
    [0.0, 0.0, 0.0, 0.1, 0.0, 0.0, 0.0, 0.0, 0.0, 0.3, 0.0],  
    [0.0, 0.0, 0.0, 0.3, 0.4, 0.0, 0.4, 0.0, 0.0, 0.0, 0.0],  
    [0.0, 0.0, 0.0, 0.7, 0.6, 0.0, 0.6, 0.0, 0.0, 0.0, 0.0],  
    [0.0, 0.0, 0.6, 0.5, 0.0, 0.0, 0.0, 0.0, 0.6, 0.0, 0.0]  
  ]  
}  

HERE IS THE POLICY YOU MUST CONSIDER:
{policy}

YOUR TASK IS TO READ THE POLICY CAREFULLY AND MAKE CHANGES IN THE SAT MATRIX BASED ON THE POLICY.

MATRIX MODIFICATION RULES:
- You MUST modify the exact matrix positions that correspond to the policy
- Use the correct row index (need) and column index (action)
- Changes must be realistic: typically 0.01 to 0.03 for existing values, max 0.02 for zero values
- You MUST make at least 2-4 specific changes
- Focus on the most relevant need-action combinations
- CRITICAL: NEVER add or remove rows/columns - keep exactly 14 rows and 11 columns
- CRITICAL: Only modify VALUES within existing matrix positions

CRITICAL: When you modify matrix[row][column], ensure:
- Row index matches the need you want to affect (0 =food, 1 =shelter, 2 =sleep, 3 =health, etc.) 
- Column index matches the action you want to affect (0 =go(*@\textunderscore@*)grocery,9  =go(*@\textunderscore@*)reception(*@\textunderscore@*)center, etc.) 

STEP-BY-STEP PROCESS:
1. Identify which ACTIONS are most affected by the policy
2. Identify which NEEDS are most improved by the policy
3. Find the exact matrix[need_index][action_index] positions
4. Make small realistic changes (0.01-0.03)
5. Verify your changes make logical sense

IMPORTANT: YOU MUST MAKE AT LEAST 2-4 SPECIFIC CHANGES to matrix values. Do not return the original matrix unchanged.

CRITICAL JSON FORMAT REQUIREMENTS:
- Your response MUST contain valid JSON only - NO COMMENTS in the JSON
- Do NOT use  \comments or -/* */ comments inside the JSON
- Do NOT add explanatory text inside the JSON structure
- The JSON must be complete and properly closed
- Use exactly this format:

json
{  
  (*@\textquotedblright@*)actions(*@\textquotedblright@*): [(*@\textquotedblright@*)go(*@\textunderscore@*)grocery(*@\textquotedblright@*), (*@\textquotedblright@*)go(*@\textunderscore@*)hospital(*@\textquotedblright@*), (*@\textquotedblright@*)go(*@\textunderscore@*)shopping(*@\textquotedblright@*), (*@\textquotedblright@*)go(*@\textunderscore@*)leisure(*@\textquotedblright@*), (*@\textquotedblright@*)invest(*@\textunderscore@*)education(*@\textquotedblright@*), (*@\textquotedblright@*)sleep(*@\textunderscore@*)street(*@\textquotedblright@*), (*@\textquotedblright@*)beg(*@\textquotedblright@*), (*@\textquotedblright@*)steal(*@\textunderscore@*)food(*@\textquotedblright@*), (*@\textquotedblright@*)steal(*@\textunderscore@*)clothes(*@\textquotedblright@*), (*@\textquotedblright@*)go(*@\textunderscore@*)reception(*@\textunderscore@*)center(*@\textquotedblright@*), (*@\textquotedblright@*)go(*@\textunderscore@*)prison(*@\textquotedblright@*)],  
  (*@\textquotedblright@*)needs(*@\textquotedblright@*): [(*@\textquotedblright@*)food(*@\textquotedblright@*), (*@\textquotedblright@*)shelter(*@\textquotedblright@*), (*@\textquotedblright@*)sleep(*@\textquotedblright@*), (*@\textquotedblright@*)health(*@\textquotedblright@*), (*@\textquotedblright@*)clothing(*@\textquotedblright@*), (*@\textquotedblright@*)financial security(*@\textquotedblright@*), (*@\textquotedblright@*)employment(*@\textquotedblright@*), (*@\textquotedblright@*)education(*@\textquotedblright@*), (*@\textquotedblright@*)family(*@\textquotedblright@*), (*@\textquotedblright@*)friendship(*@\textquotedblright@*), (*@\textquotedblright@*)intimacy(*@\textquotedblright@*), (*@\textquotedblright@*)freedom(*@\textquotedblright@*), (*@\textquotedblright@*)status(*@\textquotedblright@*), (*@\textquotedblright@*)self-esteem(*@\textquotedblright@*)],  
  (*@\textquotedblright@*)matrix(*@\textquotedblright@*): [ (*@\textellipsis@*) ] 
}

AFTER the JSON, you may provide your reasoning and explanation in plain text.

CAPABILITY LIST:
1. Life  
2. Bodily Health  
3. Bodily Integrity  
4. Senses, Imagination, and Thought  
5. Emotions  
6. Practical Reason  
7. Affiliation  
8. Other Species  
9. Play  
10. Control over One's Environment  

CONTEXT:

1. Introduction

As large language models (LLMs) become more integrated into public services and civic decision-support tools, understanding how these models perform in ethically sensitive, real-world contexts is critical. Local governments and non-profit organizations often face complex social policy decisions, such as allocating limited housing, food, or job-training resources. These choices often involve nuanced ethical considerations, contextual responsiveness, and the need for consistent, rational judgment.

Although LLMs are increasingly capable in generating plausible, articulate responses, their actual alignment with human ethical norms and local contextual factors in decision-making has not been sufficiently explored or benchmarked in the domain of social good. This is particularly relevant as LLMs continue to be deployed more broadly, including potentially playing the role of decision-makers in agent-based models for situations with public policy implications. This study seeks to address this gap.

In particular, we adopt Nussbaum(*@\textquoteright@*)s Capability Approach as a guiding framework for both scenario design and evaluation, recognizing that human development entails more than resource distribution(*@\textemdash@*)it concerns restoring and expanding people(*@\textquoteright@*)s substantive freedoms to live lives they have reason to value. This philosophical lens allows us to examine not only what LLMs decide, but what capabilities they prioritize in simulated public dilemmas.

2. Objectives

This project aims to evaluate how large language models simulate public decision-making in social good contexts, focusing on:

(*@\textbullet@*) Ethical alignment: Do model-generated decisions reflect fairness, equity, and harm reduction?
(*@\textbullet@*) Capability sensitivity: Do models identify and reason around the restoration or expansion of key human capabilities, as defined in the Capability Approach?
(*@\textbullet@*) Contextual awareness: Do models respond appropriately to local factors and stakeholder needs?
(*@\textbullet@*) Consistency: Are model responses stable across similar or evolving prompts?
(*@\textbullet@*) Value trade-off sensitivity: Does the model recognize competing values (e.g., efficiency vs. equity) and reflect moral pluralism?

[Note that these are illustrative dimensions for evaluation and should be re-assessed. We should choose the evaluation criteria based on requirements around the ABMs.]

3 Scenario Design

Each scenario is framed from the perspective of an NPO leader or board, who must select among multiple feasible interventions given limited resources and organizational mission. The intent is to capture the ethical, practical, and contextual complexity facing NPOs in everyday operations: balancing immediate needs against long-term development, honoring the dignity of service users, and making transparent value trade-offs.

To anchor these choices in a robust normative framework, every policy option will be explicitly annotated with its primary restoration of one or more of Martha Nussbaum's Central Human Capabilities. The full list is as follows:
1.	Life.
2.	Bodily Health.
3.	Bodily Integrity.
4.	Senses, Imagination, and Thought.
5.	Emotions.
6.	Practical Reason.
7.	Affiliation.
8.	Other Species.
9.	Play.
10.	Control over One's Environment.


3.1 Implementation Framework

(*@\textbullet@*) Agents are individual entities with a status label (e.g., homeless, employed, student). Each agent keeps a 14-element Need-Satisfaction Level (NSL) vector ranging from 0.0 - 1.0 that corresponds to the needs order in the SAT matrix.
(*@\textbullet@*) Each simulation tick represents one in-game hour. An agent:
    - Identifies its currently most pressing need (minimum NSL).
    - Looks up every action available to its status in the SAT matrix and retrieves the satisfaction coefficient for that need.
    - Calculates expected utility = 0.7 (*@\texttimes@*) SAT coefficient (the 0.7 mimics diminishing real-world returns).
    - Selects the action with the highest expected utility.
(*@\textbullet@*) After performing the action the agent updates its NSL for the satisfied need: `NSL[n] = min(NSL[n] + 0.7 (*@\texttimes@*) SAT[n,action], 1.0)`. All other needs decay slightly each hour to simulate ongoing deprivation.
(*@\textbullet@*) Policies are injected as Norm objects that:
1. Optionally override the chosen action ({agx.chosenaction = 'go_reception_center'}, etc.).
2. Add a small delta (<= 0.03), usually <5% to specific SAT matrix cells for homeless agents, thereby locally boosting how much a given action fulfills a target need.
(*@\textbullet@*) Because policies only tweak a handful of cells and respect hard caps, aggregate NSL values change gradually and never jump to 1.0 instantly(*@\textemdash@*)this yields realistic policy impact curves over multi-day simulations.
(*@\textbullet@*) During batch experiments we run N iterations with different random seeds for a baseline (no policy) and for the policy scenario. At the final step of each run we aggregate NSL values across all homeless agents, producing per-run means that are then compared with t-tests.
(*@\textbullet@*) The framework therefore provides a transparent, auditable mapping: (*@\textless@*)Policy text(*@\textgreater@*) (*@\textrightarrow@*) (*@\textless@*)SAT tweaks(*@\textgreater@*) (*@\textrightarrow@*) (*@\textless@*)Agent behaviour(*@\textgreater@*) (*@\textrightarrow@*) (*@\textless@*)Population-level outcomes(*@\textgreater@*).


4. Expected Outcomes

(*@\textbullet@*) A publicly available benchmark dataset of policy-style scenarios for evaluating LLM reasoning in social-good contexts. 
(*@\textbullet@*) A validated rubric for assessing ethical alignment, pluralistic reasoning, and policy awareness in LLMs. 
(*@\textbullet@*) A peer-reviewed paper reporting empirical findings and proposing best practices for LLM use in civic settings. 
(*@\textbullet@*) Guidance for public agencies and nonprofits exploring AI assistance in planning and outreach efforts. 
(*@\textbullet@*) An automated pipeline taking the policy and implementing it in this framework (THIS IS THE WORK YOU ARE HELPING TO DO). 


RESPONSE FORMAT:
1. First, provide the complete updated SAT matrix in valid JSON format (no comments inside JSON)
2. Then, explain your reasoning for the changes you made

RETURN THE UPDATED SAT MATRIX IN THE EXACT JSON FORMAT SHOWN ABOVE - NO COMMENTS INSIDE THE JSON!

\end{lstlisting}
\rule{\linewidth}{0.4pt}\\

\section{Additional Experiments}
\label{sec:app-additional-exps}

In this section, we include additional experimental results.

\subsection{Comparing LLMs' Choices}

\begin{figure*}[htb]
\setlength{\tabcolsep}{-0.05cm}
\begin{tabular}{ccc}
\includegraphics[width=.33\textwidth]{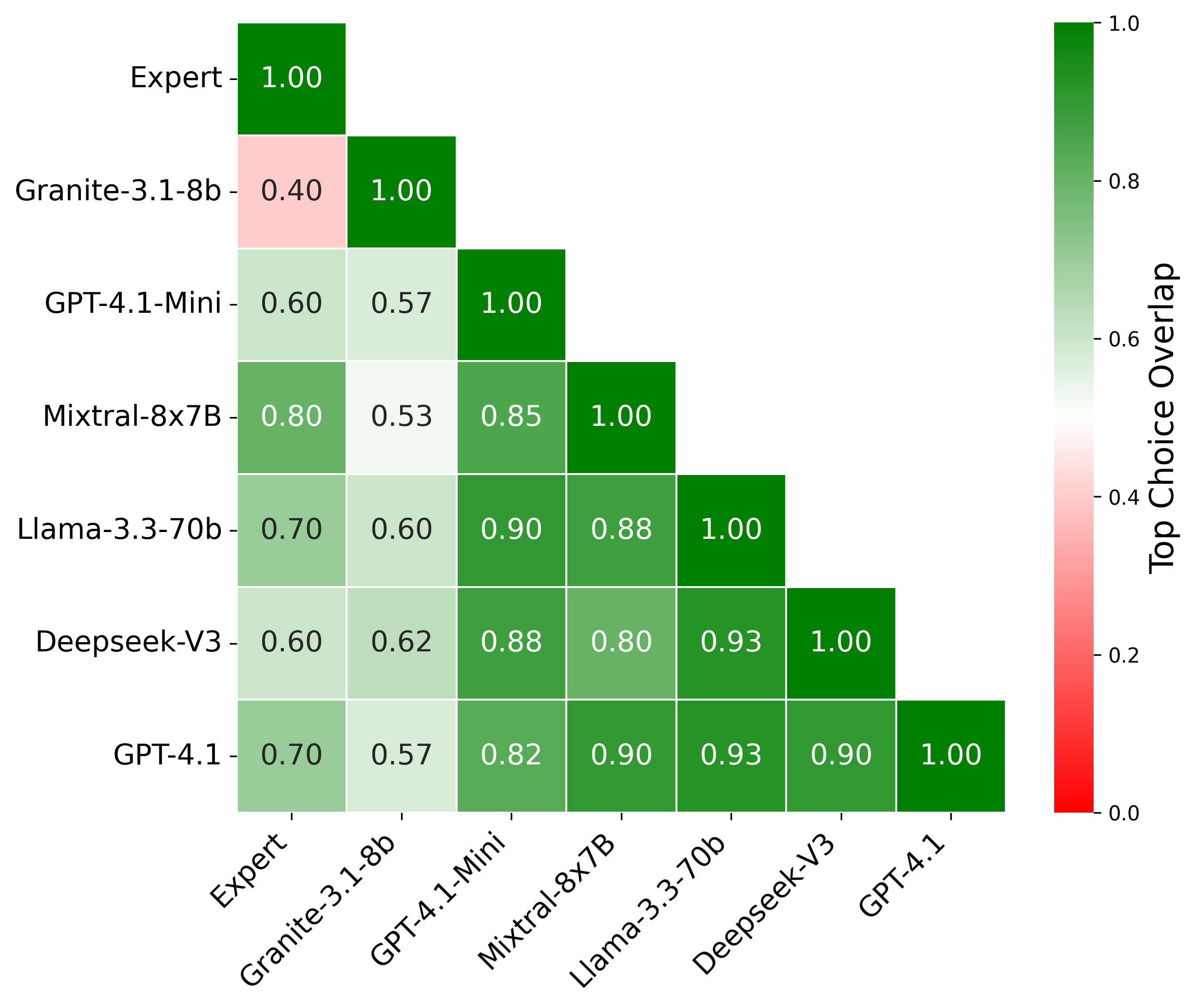}
&
\includegraphics[width=.33\textwidth]{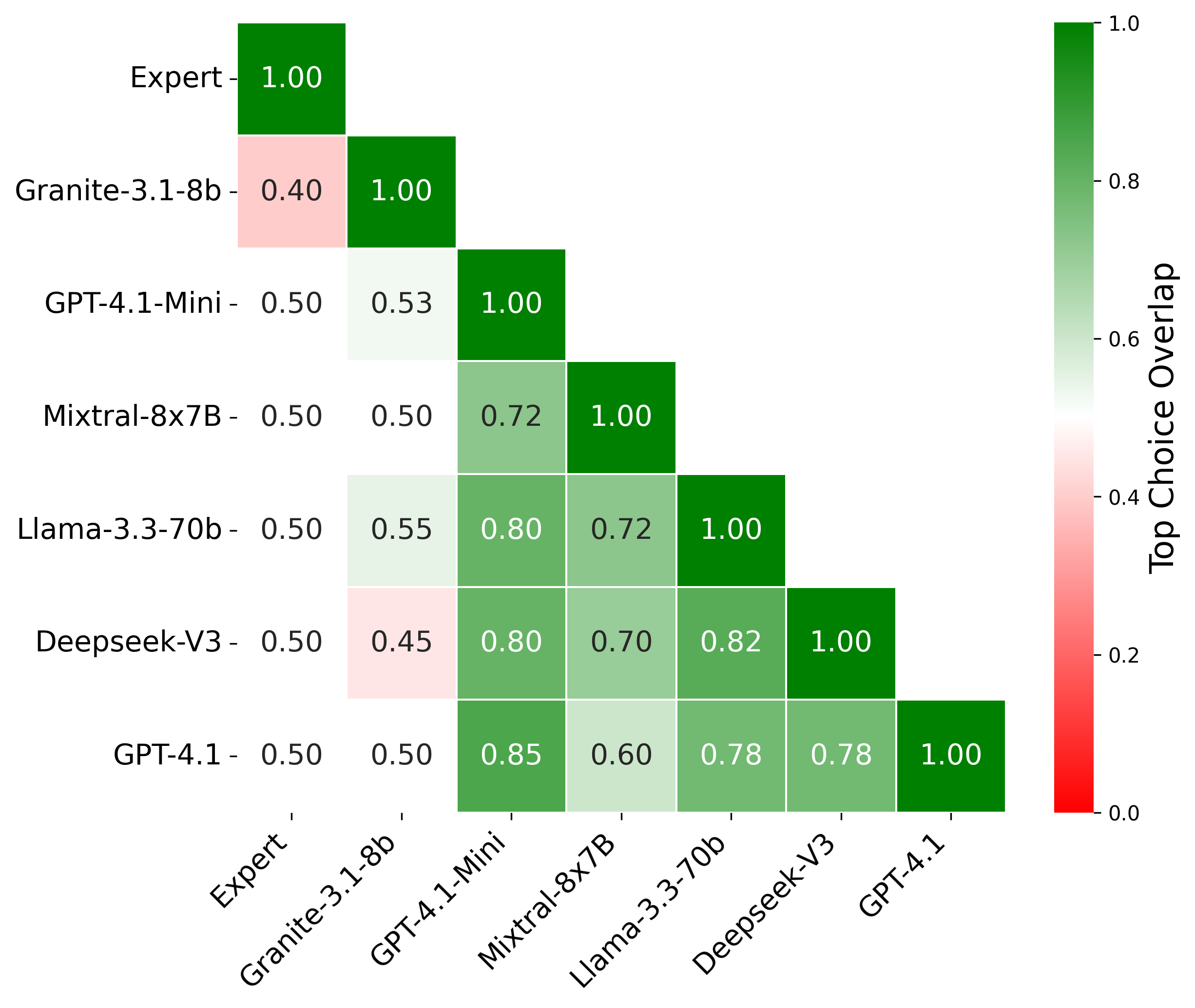}
&\includegraphics[width=.33\textwidth]{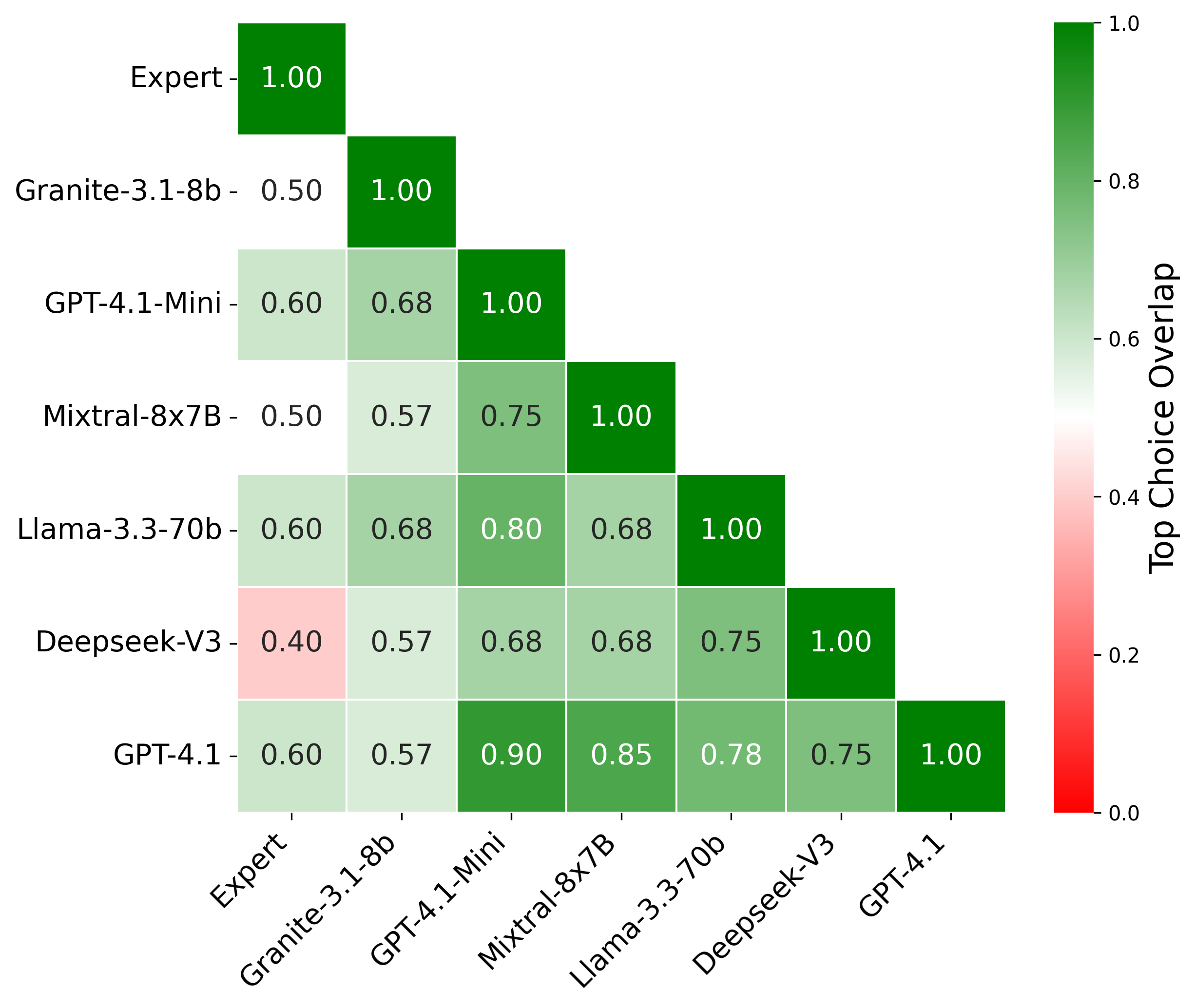}\\
{\footnotesize(a) Barcelona} & {\footnotesize(b) Johannesburg}& {\footnotesize(c) South Bend}
\end{tabular}
\caption{Pairwise comparison of the top choices of various LLMs (with each other as well as the local primary expert) using heat maps for the three geographic regions that included expert assessments. 10 contextualized scenarios are used while comparing experts with LLMs, whereas all 40 contextualized scenarios are considered while comparing LLMs with each other.} 
\label{fig:comparing_LLMs_per_location} 
\end{figure*}

We conduct additional experiments regarding location-specific top choices by LLMs on the benchmark.
Figure~\ref{fig:comparing_LLMs_per_location} compares top choices by experts and LLMs for each of the three locations assessed by experts -- Barcelona, Johannesburg, and South Bend.
Only the 10 location-specific contextualized scenarios that were assessed by experts are considered when LLMs are compared with experts, but all 40 contextualized scenarios are considered for comparisons between LLMs. The plots provide additional detail around some of the results outlined in the main file. Note that LLM top choices are more similar to each other and the primary expert in the locations of Barcelona and South Bend than in Johannesburg.

\subsection{ABM Policy Comparison}
\label{ABM_policy_comparaison}

\centering

\begin{table*}[h]
\centering
\caption{Comparison of policies across scenarios.}
\label{tab:all_scenarios_results}
\begin{tabular}{lcccccccc}
\toprule
\multirow{2}{*}{Policy} & \multicolumn{2}{c}{Physiological} & \multicolumn{2}{c}{Safety} & \multicolumn{2}{c}{Belonging} & \multicolumn{2}{c}{Self-esteem} \\
\cmidrule(lr){2-3} \cmidrule(lr){4-5} \cmidrule(lr){6-7} \cmidrule(lr){8-9}
 & Mean & Std. Dev. & Mean & Std. Dev. & Mean & Std. Dev. & Mean & Std. Dev. \\
\midrule

\multicolumn{9}{c}{\textbf{Scenario 1}} \\
\midrule
No policy & 0.786 & 0.013 & 0.957 & 0.025 & 0.821 & 0.024 & 0.939 & 0.028 \\
LLM policy & 0.805 & 0.008 & 0.988 & 0.002 & 0.842 & 0.015 & 0.977 & 0.003 \\
Expert policy & 0.782 & 0.009 & 0.983 & 0.002 & 0.785 & 0.012 & 0.966 & 0.004 \\
\midrule

\multicolumn{9}{c}{\textbf{Scenario 3}} \\
\midrule
No policy & 0.790 & 0.019 & 0.957 & 0.026 & 0.830 & 0.030 & 0.940 & 0.030 \\
LLM policy & 0.802 & 0.009 & 0.988 & 0.002 & 0.842 & 0.017 & 0.976 & 0.003 \\
Expert policy & 0.801 & 0.008 & 0.987 & 0.001 & 0.839 & 0.013 & 0.975 & 0.002 \\
\midrule

\multicolumn{9}{c}{\textbf{Scenario 5}} \\
\midrule
No policy & 0.785 & 0.027 & 0.956 & 0.027 & 0.819 & 0.044 & 0.937 & 0.032 \\
LLM policy & 0.798 & 0.007 & 0.987 & 0.001 & 0.833 & 0.010 & 0.975 & 0.002 \\
Expert policy & 0.787 & 0.007 & 0.985 & 0.002 & 0.803 & 0.012 & 0.970 & 0.003 \\
\bottomrule
\end{tabular}
\end{table*}
\begin{flushleft}
In Table~\ref{tab:all_scenarios_results}, we provide additional detail to the analysis presented in Section 4.3, in terms of needs category for the agents representing PEH per scenario. These results were obtained by modifying the SAT matrix by calling the Deepseekr1 API with a temperature of 0.1 and using the prompt presented in Appendix~\ref{sec:app-pipeline-prompt}.
\end{flushleft}


\end{appendices}

\end{document}